\documentclass[10pt,twocolumn,letterpaper]{article}
\usepackage[pagenumbers]{cvpr}
\usepackage{microtype}
\usepackage{mathtools}
\usepackage{amsthm}
\usepackage{algorithm}
\usepackage{algpseudocode}
\usepackage{tabularx}
\usepackage{booktabs}
\usepackage{multirow}
\newcommand{\method}{RetroCBM}

\newcommand{\R}{\mathbb R}

\newcommand{\Emp}{\widehat{\mathbb E}_t}

\DeclareMathOperator*{\argmin}{arg\,min}
\DeclareMathOperator*{\argmax}{arg\,max}
\newtheorem{proposition}{Proposition}
\usepackage{tikz}
\usetikzlibrary{positioning,arrows.meta,fit,calc,backgrounds,decorations.pathreplacing,shapes.geometric}

\definecolor{bpOurs}{RGB}{0,114,178}      %
\definecolor{bpPrior}{RGB}{213,94,0}      %
\definecolor{bpData}{RGB}{0,158,115}      %
\definecolor{bpAccent}{RGB}{230,159,0}    %
\definecolor{bpGray}{RGB}{110,110,110}    %
\definecolor{bpLight}{RGB}{238,238,238}   %
\colorlet{bpOursFill}{bpOurs!18}
\colorlet{bpPriorFill}{bpPrior!15}
\colorlet{bpDataFill}{bpData!15}

\tikzset{
  bp/base/.style   ={draw=black!70, line width=0.5pt, rounded corners=2pt, align=center,
                     font=\sffamily\scriptsize, minimum height=0.6cm, inner sep=3pt},
  bp/module/.style ={bp/base, fill=white},
  bp/frozen/.style ={bp/base, fill=bpLight, draw=bpGray, text=black!80},
  bp/ours/.style   ={bp/base, fill=bpOursFill, draw=bpOurs, line width=0.7pt},
  bp/prior/.style  ={bp/base, fill=bpPriorFill, draw=bpPrior},
  bp/data/.style   ={bp/base, fill=bpDataFill, draw=bpData},
  bp/todo/.style   ={bp/base, dashed, draw=bpGray, text=bpGray},
  bp/flow/.style   ={-{Latex[length=1.6mm]}, line width=0.6pt, black!75},
  bp/note/.style   ={font=\sffamily\tiny, text=bpGray, align=center},
  bp/panel/.style  ={font=\sffamily\footnotesize\bfseries, anchor=west},
  bp/frame/.style  ={draw=black!70, fill=white, line width=0.4pt},
}

\pdftrailerid{}
\definecolor{cvprblue}{rgb}{0.21,0.49,0.74}
\usepackage[pagebackref,breaklinks,colorlinks,allcolors=cvprblue]{hyperref}
\def\confName{CVPR}
\def\confYear{2027}
\title{Retroactive Concept Bottleneck Models for Exemplar-Free \\ Class-Incremental Learning}
\author{
Yang Ji$^{1}$ \quad Ying Sun$^{2}$ \quad Yuanxin Zhuang$^{1}$ \quad Yuting Zhang$^{1}$\\
{$^{1}$The Hong Kong University of Science and Technology (Guangzhou)}\\
{$^{2}$The 63rd Research Institute, National University of Defense Technology}\\
{\tt\small yji655@connect.hkust-gz.edu.cn \quad sunying@nudt.edu.cn}\\
{\tt\small yzhuang436@connect.hkust-gz.edu.cn \quad yzhang755@connect.hkust-gz.edu.cn}
}
\begin{document}
\maketitle
\begin{abstract}
Exemplar-free class-incremental learning (EFCIL) requires models to learn new classes sequentially without retaining training samples from previous tasks.
Continual concept bottleneck models (CBMs) make their predictions interpretable by identifying visual concepts in an image and using them to score each class.
However, these methods typically focus on preserving learned concepts or approximating past classes with generated features, which can leave new concepts underused in old-class predictions.
We propose RetroCBM, which retains a small set of vocabulary-independent statistics relating frozen visual features to the embedding space of a pretrained vision-language model.
Even after earlier training images have been discarded, we can use these statistics to supervise new concepts. To do so, we develop a retroactive regression rule that yields the same predictor for each new concept as class-balanced ridge regression on all images seen so far.
We then reproject the stored statistics to recover the full-history squared classification loss and thus learn how much each new concept should contribute to old-class predictions. Since the classifier may use systematic concept errors to distinguish nearby classes, we further introduce a graph-weighted penalty on the margin shifts caused by those errors.
Across six datasets, RetroCBM outperforms state-of-the-art interpretable methods in all class-incremental settings, with an 18.1\% relative gain in mean average incremental accuracy over the strongest exemplar-free baseline.
\end{abstract}

\section{Introduction}
\label{sec:intro}
In many visual recognition applications, data of new categories would arrive after the initial model has been trained.
Class-incremental learning (CIL) addresses this setting by training a model on successive groups of classes while requiring it to distinguish among all classes seen to date~\cite{zhang2023slca,mcdonnell2023ranpac}.
The exemplar-free setting (EFCIL) is of particular interest when storage or privacy constraints prevent the model from retaining earlier training images~\cite{zhuang2022acil,petit2023fetril}.
To keep predictions accurate on earlier classes, existing methods constrain changes to visual representations during fine-tuning~\cite{zhu2021pass,zhu2022ssre} or update the classifier over a fixed set of features~\cite{petit2023fetril,goswami2023fecam}.
How the model arrives at those predictions, however, has received less attention~\cite{rymarczyk2023icicle,yu2025clgcbm,javadi2026cicbm}.
Can a model maintain accuracy across old and new classes while making clear which visual attributes support each decision?

\begin{figure}[t]
    \centering
    \includegraphics[width=\columnwidth]{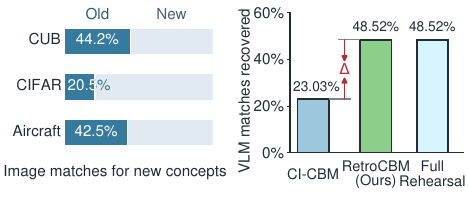}
    \vspace{-6mm}
    \caption{Images from earlier classes often exhibit attributes named only in a later phase. Left, the share of VLM matches for second-phase concepts that come from earlier classes. Right, the fraction of old-image matches recovered on CUB.}
    \vspace{-3mm}
    \label{fig:motivation}
\end{figure}

Concept bottleneck models (CBMs) offer a promising approach to interpretable CIL by basing class predictions on recognizable visual attributes~\cite{koh2020cbm,11275698}.
A CBM trained with supervision from a pretrained vision-language model (VLM) predicts which concepts are present in an image and uses those predictions to score each candidate class~\cite{radford2021clip,oikarinen2023labelfree}.
In continual learning, CBMs must balance stability and plasticity as new concepts are introduced.
To preserve old concepts, existing methods typically regularize concept responses~\cite{rymarczyk2023icicle} or freeze previously learned concept mappings~\cite{yang2024conceptdriven}.
To use new concepts when recognizing earlier classes, other methods generate synthetic features for those classes and use the resulting concept scores to update the classifier~\cite{javadi2026cicbm,yu2025clgcbm}.
However, by focusing on how classifier uses new concepts, these methods typically overlook earlier images as a source of supervision for learning those concepts. Earlier training images may show the relevant attributes long before the concepts are introduced.
For example, if \emph{webbed feet} is introduced only when \emph{geese} arrive, the \emph{duck} images that could have helped the model learn to recognize this attribute have already been discarded.
Across the three datasets in \cref{fig:motivation} (left), images from earlier classes account for an average of 35.7\% of the VLM matches for concepts introduced later.
On CUB (right), CI-CBM~\cite{javadi2026cicbm} recovers 23.0\% of the VLM's old-image matches, while RetroCBM and full rehearsal each recover 48.5\%.
These findings suggest that learning new concepts from earlier images could improve old-class recognition and support explanations for all seen classes in a shared, expanding concept space.

However, applying new concepts to previously learned classes poses three key challenges.
First, as new classes introduce new concepts, the model can no longer examine earlier training images to determine whether and to what extent they exhibit those concepts.
Second, even after the model has learned a new concept, it must determine how to use it alongside existing concepts when recognizing earlier classes.
For example, webbed feet can help distinguish ducks from birds that lack them, but offer little help in telling ducks and geese apart.
Third, it is difficult to assess how much a decision relies on errors in concept prediction.
If equally visible webbed feet consistently receive lower scores in ducks than in geese, the model may use that bias to separate them.

We introduce RetroCBM, a label-free concept bottleneck model for EFCIL with a shared, expanding concept vocabulary. Our central observation is that a small set of statistics collected from earlier images can provide supervision for concepts named later. Before discarding the images, RetroCBM records how their visual features relate to the image embeddings that the VLM uses to score concept descriptions. When a new description arrives, retroactive concept regression uses these statistics to recover the same concept predictor as class-balanced ridge regression on all images seen so far. We then propose moment reprojection to recover the classification loss in the updated concept space, allowing each earlier class to learn which new concepts help distinguish it from other classes. To limit the effect of concept prediction errors, our graph-weighted margin distortion penalty discourages changes in the score margin between similar classes when predicted concept scores replace the VLM scores. Our key contributions are as follows.
\begin{itemize}
    \item We find that images from earlier classes often contain attributes named only later. On CUB, human annotations show that supervision from these images improves recognition of newly named color attributes.
    \item We introduce \method{}, a label-free CBM that uses stored statistics to recover full-history ridge predictors for new concepts and refit old-class classifiers in the expanded concept space.
    \item Across six datasets, \method{} achieves the highest average incremental accuracy (AIA) in all 16 settings, with an 18.1\% relative gain in mean AIA over CI-CBM, the strongest exemplar-free interpretable method.
\end{itemize}

\begin{table}[t]
    \centering
    \caption{\textbf{Comparison with related methods.}
    \emph{Retroactive} and \emph{Update} denote using old-class evidence
    to learn new concepts and their old-class weights, respectively.
    \emph{Equivalence} means exactly matching the full-history solution or loss.
    $\approx$ denotes exemplar replay~\cite{agrawal2025mucil} or
    pseudo-features~\cite{yu2025clgcbm,javadi2026cicbm}. A dash means inapplicable.}
    \vspace{-2mm}
    \label{tab:method-comparison}
    \small
    \setlength{\tabcolsep}{2pt}
    \begin{tabular*}{\columnwidth}{@{\extracolsep{\fill}}lcccc@{}}
        \toprule
        Method & Interpretability & Retroactive & Update & Equivalence \\
        \midrule
        \cite{zhuang2022acil,mcdonnell2023ranpac}
            & $\times$ & -- & -- & $\checkmark$ \\
        \cite{rymarczyk2023icicle,yang2024conceptdriven}
            & $\checkmark$ & $\times$ & $\times$ & $\times$ \\
        \cite{agrawal2025mucil,yu2025clgcbm}
            & $\checkmark$ & $\approx$ & $\approx$ & $\times$ \\
        \cite{lai2025concil}
            & $\checkmark$ & $\times$ & $\times$ & $\checkmark$ \\
        \cite{javadi2026cicbm}
            & $\checkmark$ & $\times$ & $\approx$ & $\times$ \\
        \midrule
        \textbf{Ours}
            & $\checkmark$ & $\checkmark$ & $\checkmark$ & $\checkmark$ \\
        \bottomrule
    \end{tabular*}
    \vspace{-4mm}
\end{table}

\section{Related Work}
\label{sec:related}
\subsection{Exemplar-Free Class-Incremental Learning}
\label{sec:rlw-efcil}
Class-incremental learning (CIL) aims to learn new classes sequentially while recognizing all seen classes without task identifiers at test time.
Exemplar-free CIL further prohibits retaining historical training examples.
Its central challenge is to acquire new knowledge while preserving old-class discrimination without revisiting previous data.
Existing approaches address this challenge through complementary strategies.
Distillation and regularization constrain changes to learned representations~\cite{zhu2021pass,zhu2022ssre}, while selective parameter adaptation and architectural expansion accommodate new knowledge with reduced interference~\cite{zhang2023slca,zhou2024ease,zhao2024safe}.
Some methods summarize earlier classes in feature space~\cite{petit2023fetril,goswami2023fecam}. ACIL~\cite{zhuang2022acil}, RanPAC~\cite{mcdonnell2023ranpac}, and DPCR~\cite{he2025dpcr} use historical feature statistics to update class predictors, with DPCR also accounting for feature drift. 
By contrast, \method{} uses statistics from earlier images to supervise concepts named only after those images have been discarded.

\subsection{Concept Bottleneck Models}
\label{sec:rlw-cbm}
Concept bottleneck models~\cite{koh2020cbm} use human-interpretable concepts as an intermediate representation, enabling users to understand model decisions and intervene on concept predictions.
Post-hoc CBMs~\cite{yuksekgonul2023posthoc} construct concept predictors from pre-trained representations, while LF-CBM~\cite{oikarinen2023labelfree} and LaBo~\cite{yang2023labo} use foundation models to reduce dependence on concept annotations.
In continual learning, the concept vocabulary expands as new classes arrive.
ICICLE~\cite{rymarczyk2023icicle} and IN2~\cite{yang2024conceptdriven} preserve existing concepts by regularizing visual prototypes and freezing existing concept mappings, respectively.
CONCIL~\cite{lai2025concil} recursively updates its concept and class predictors from each phase's labels and representations.
CLG-CBM~\cite{yu2025clgcbm} and CI-CBM~\cite{javadi2026cicbm} use pseudo-features to represent earlier classes. CI-CBM projects these pseudo-features through the current concept layer to learn how new concepts contribute to old-class predictions. As \cref{tab:method-comparison} shows, a new concept can contribute to old-class predictions even when its predictor was learned without supervision from earlier classes.
By contrast, \method{} learns these weights by reprojecting stored visual statistics into the expanded concept space.
For a fixed current concept layer, this recovers the same class-balanced squared loss that would be obtained by evaluating the classifier on all images seen so far.

\section{Preliminaries}
\label{sec:prelim}
\textbf{Problem Setup.}
We formally define exemplar-free class-incremental learning for concept bottleneck models with an expanding concept layer and class prediction space.
At phase $t\in\{1,\ldots,T\}$, the learner receives a dataset $\mathcal D_t=\{(x_i,y_i)\}_{i=1}^{N_t}$ of $N_t$ labeled images from new classes $\Delta\mathcal Y_t$ and a set of new concept descriptions $\Delta\mathcal C_t$.
Starting from $\mathcal Y_0=\mathcal C_0=\varnothing$, the cumulative sets become $\mathcal Y_t=\mathcal Y_{t-1}\cup\Delta\mathcal Y_t$ and $\mathcal C_t=\mathcal C_{t-1}\cup\Delta\mathcal C_t$.
Their sizes, $K_t=|\mathcal Y_t|$ and $M_t=|\mathcal C_t|$, denote the numbers of seen classes and concepts, respectively.
Classes are disjoint across phases, whereas concepts may also describe images from earlier classes.
After each phase, the learner discards its training images and their individual feature vectors. It may retain aggregate statistics across phases.
At test time, the model predicts among all $K_t$ seen classes without task identifiers.

\method{} uses a linear concept layer followed by a linear classifier.
Let $\mathbf z=f(x)\in\R^d$ denote the features extracted from image $x$ by a frozen visual encoder.
At phase $t$, the two layers are parameterized as
\begin{equation}
 \begin{aligned}
 \hat{\mathbf c}^{(t)}(x)&=\mathbf W_c^{(t)}\mathbf z,\\
 \mathbf s^{(t)}(x)&=\mathbf W_y^{(t)}\hat{\mathbf c}^{(t)}(x)+\mathbf b^{(t)}.
 \end{aligned}
 \label{eq:model}
\end{equation}
Here, $\mathbf W_c^{(t)}\in\R^{M_t\times d}$ projects visual features into the concept space.
The classification layer uses weights $\mathbf W_y^{(t)}\in\R^{K_t\times M_t}$ and bias $\mathbf b^{(t)}\in\R^{K_t}$ to produce class scores.
Each response $\hat c_j^{(t)}(x)$ corresponds to a concept $c_j\in\mathcal C_t$, and the final prediction is $\hat y=\argmax_{k\le K_t}s_k^{(t)}(x)$.
Following prior work~\cite{oikarinen2023labelfree,javadi2026cicbm}, we supervise concept predictions with image--text similarity scores from a frozen vision-language model (VLM).

\noindent\textbf{Learning objectives.}
At each phase, learning must balance two objectives:
\textit{(1) Stability:} keep predictions for existing concepts aligned with their VLM reference scores and continue to recognize earlier classes among all seen classes.
\textit{(2) Plasticity:} learn new concepts and classes and use the new concepts to recognize earlier classes.
Once earlier images have been discarded, the learner can no longer score them against new concept descriptions or pass them through the updated concept layer to update the classifier.
Concept prediction errors can also alter the score margin between two classes. To tackle these challenges, we thus introduce retroactive concept regression, moment reprojection, and a graph-weighted margin distortion penalty in next section.

\section{Methodology}
\label{sec:method}
\begin{figure*}[t]
    \centering
    \includegraphics[width=0.95\linewidth]{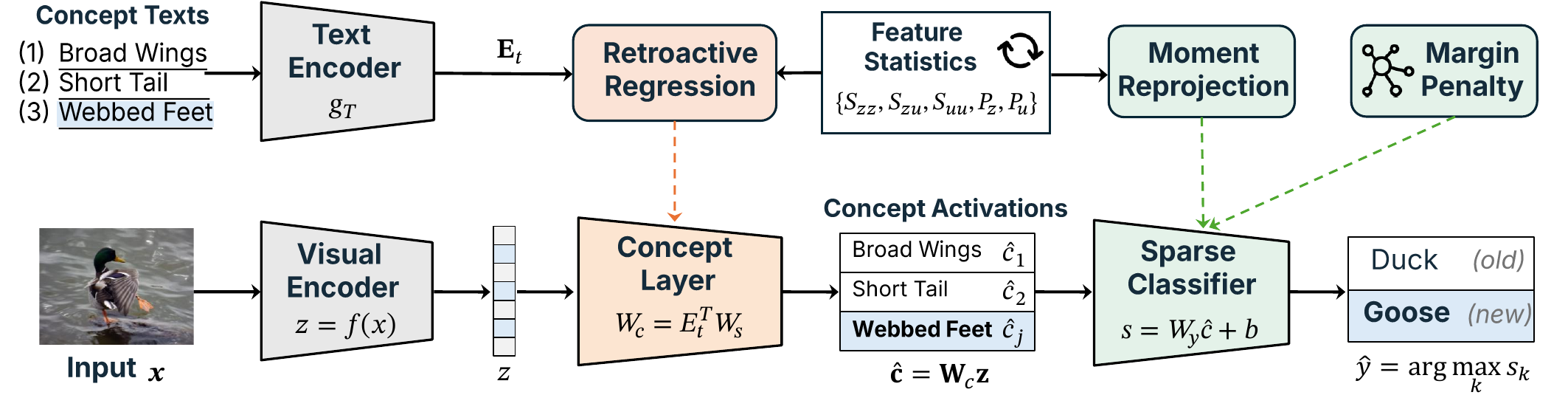}
    \vspace{-1mm}
    \caption{\textbf{Model architecture.} At inference (bottom), the sparse classifier scores all seen classes from the concept responses predicted for each image. During training (top), retroactive regression updates all current concept predictors using stored statistics from all seen classes. We then reproject these statistics to update the classifier and penalize error-induced margin shifts between neighboring classes.}
    \vspace{-3mm}
    \label{fig:overview}
\end{figure*}

Suppose the concept \emph{webbed feet} is introduced with \emph{geese} after the training images of \emph{ducks} have been discarded.
As shown in \cref{fig:overview}, \method{} uses stored feature statistics to fit concept predictors to VLM reference scores across all seen classes, so the earlier images of \emph{ducks} still contribute to learning the new concept (\cref{sec:concept-learning}).
With the concept layer fixed, we reproject the stored visual statistics to update the classifier in the expanded concept space, allowing it to learn how much \emph{webbed feet} should contribute to the score for \emph{ducks} (\cref{sec:reproject}).
During this update, we penalize shifts in the margins between neighboring classes when predicted concept scores replace the VLM reference scores (\cref{sec:error}).

\subsection{Retroactive Concept Regression}
\label{sec:concept-learning}
At phase $t$, we need to introduce new concepts for the visual attributes of the new classes.
Earlier classes may share some of these attributes, but their training images are no longer available.
Our goal is to recover the concept layer we would fit by scoring every image seen so far against every concept in $\mathcal C_t$.
Here, we use image-text matching scores from a pretrained VLM to supervise concept prediction without manual concept annotations.
The visual encoder and both VLM encoders remain frozen across phases.
Let $\mathbf u=g_I(x)\in\R^m$ be the VLM image embedding and let the columns of $\mathbf E_t=[\mathbf e_1,\ldots,\mathbf e_{M_t}]$ be the text embeddings $\mathbf e_j=g_T(c_j)$.
Each embedding has unit norm, so the reference scores $\mathbf c_{\mathrm{ref}}^{(t)}(x)=\mathbf E_t^\top\mathbf u$ are image-text cosine similarities.
We formulate concept learning as a class-balanced ridge regression from frozen visual features to VLM reference scores over all images seen through phase $t$:
\begin{equation}
 \mathbf W_c^{(t)}=\argmin_{\mathbf W}\;
 \Emp\!\left[\|\mathbf W\mathbf z-\mathbf E_t^\top\mathbf u\|_2^2\right]
 +\lambda_s\|\mathbf W\|_F^2 .
 \label{eq:concept-ridge}
\end{equation}
Here $\Emp$ gives each seen class equal weight. For a class $k$ with $n_k$ images, this amounts to a weight of $1/(K_t n_k)$ per image.
For $\lambda_s>0$, the unique solution of \cref{eq:concept-ridge} is
\begin{equation}
 \mathbf W_c^{(t)}
 =\Emp\!\left[(\mathbf E_t^\top\mathbf u)\mathbf z^\top\right]
   \bigl(\mathbf S_{zz}^{(t)}+\lambda_s\mathbf I\bigr)^{-1},
 \label{eq:concept-solution}
\end{equation}
where $\mathbf S_{zz}^{(t)}=\Emp[\mathbf z\mathbf z^\top]$ depends only on visual features.
For each new concept, the cross-moment in \cref{eq:concept-solution} depends on its VLM scores on earlier training images. Those images are no longer available for scoring.
We call this \emph{retroactive concept regression} because statistics collected before a concept is named allow earlier images to contribute to its ridge fit.
The text embeddings in $\mathbf E_t$ are the same for every image. Taking $\mathbf E_t^\top$ outside the class-balanced average gives
\begin{equation}
 \Emp\!\left[(\mathbf E_t^\top\mathbf u)\mathbf z^\top\right]
 =\mathbf E_t^\top\Emp[\mathbf u\mathbf z^\top]
 =\mathbf E_t^\top\mathbf S_{zu}^{(t)\top}.
 \label{eq:cross-moment-factorization}
\end{equation}
We therefore accumulate the vocabulary-independent moment $\mathbf S_{zu}^{(t)}=\Emp[\mathbf z\mathbf u^\top]$ while the images are available.
For a new concept with text embedding $\mathbf e_j$, the product $\mathbf S_{zu}^{(t)}\mathbf e_j$ gives exactly the feature-score cross-moment we would obtain by scoring all the images seen through phase $t$ against that concept.

At each phase, we update $\mathbf S_{zz}$ and $\mathbf S_{zu}$ from the incoming images.
Let $\mathcal I_k$ index the $n_k$ images of class $k$.
For any pair of encoder outputs $\mathbf a,\mathbf b$ from the same image, the class-balanced moment $\mathbf S_{ab}^{(t)}=\Emp[\mathbf a\mathbf b^\top]$ satisfies
\begin{equation}
 \mathbf S_{ab}^{(t)}
 =\frac{K_{t-1}}{K_t}\mathbf S_{ab}^{(t-1)}
 +\frac{1}{K_t}\sum_{k\in\Delta\mathcal Y_t}
   \frac{1}{n_k}\sum_{i\in\mathcal I_k}\mathbf a_i\mathbf b_i^\top .
 \label{eq:moment-update}
\end{equation}
Each new class contributes its mean outer product, and every seen class has total weight $1/K_t$, as in \cref{eq:concept-ridge}.
Substituting \cref{eq:cross-moment-factorization} into the ridge solution gives
\begin{equation}
 \begin{aligned}
 \mathbf W_s^{(t)}
 &=\mathbf S_{zu}^{(t)\top}
   \bigl(\mathbf S_{zz}^{(t)}+\lambda_s\mathbf I\bigr)^{-1},\\
 \mathbf W_c^{(t)}&=\mathbf E_t^\top\mathbf W_s^{(t)} .
 \end{aligned}
 \label{eq:align-solution}
\end{equation}
The shared map $\mathbf W_s^{(t)}$ predicts the VLM image embedding $\mathbf u$ from the visual feature vector $\mathbf z$. As new classes arrive, we update this map using the accumulated statistics of all seen classes. We then compute the predictor for each concept $c_j$ in current vocabulary as $\mathbf e_j^\top\mathbf W_s^{(t)}$, the $j$th row of $\mathbf W_c^{(t)}$.

\begin{proposition}[Full-history equivalence for the current vocabulary]
\label{prop:concept-fit}
For any concept in $\mathcal C_t$, consider ridge regression
from frozen visual features to that concept's VLM scores on all images
seen through phase $t$. With the same class-balanced sample weights
and ridge coefficient $\lambda_s>0$, retroactive concept regression
recovers the same predictor exactly from the stored statistics.
\end{proposition}
Because the ridge objective in \cref{eq:concept-ridge} separates across concepts, the predictor for $c_j$ depends on the images only through $\mathbf S_{zz}^{(t)}$ and $\mathbf S_{zu}^{(t)}\mathbf e_j$. Even classes seen before $c_j$ was named contribute to this fit with weight $1/K_t$ each.

\subsection{Moment Reprojection for Classifier Updates}
\label{sec:reproject}

With the concept layer $\mathbf W_c^{(t)}$ fixed, we next learn how the new and existing concepts should contribute to each class score. For earlier classes, this requires assigning weights to concepts introduced after their images were discarded.
We formulate the classification loss over all images seen through phase $t$, using their current concept predictions $\hat{\mathbf c}^{(t)}(x)=\mathbf W_c^{(t)}\mathbf z$ and one-hot labels $\mathbf y$.
\begin{equation}
 \mathcal L_{\mathrm{cls}}^{(t)}(\mathbf W_y,\mathbf b)
 =\tfrac12\Emp\!\left[
 \|\mathbf W_y\hat{\mathbf c}^{(t)}(x)+\mathbf b-\mathbf y\|_2^2\right].
 \label{eq:class-loss}
\end{equation}
Evaluating this loss image by image would require passing earlier images through the updated concept layer. Because the concept layer is linear, we can compute the same squared loss by reprojecting the stored visual statistics into the current concept space.

When class $k$ arrives, we store its mean visual feature $\boldsymbol\mu_{z,k}$ as column $k$ of $\mathbf P_z^{(t)}$.
For any seen class $k$, linearity gives its mean response to every current concept,
\begin{equation}
 \mathbf W_c^{(t)}\boldsymbol\mu_{z,k}
 =\frac{1}{n_k}\sum_{i\in\mathcal I_k}\mathbf W_c^{(t)}\mathbf z_i .
 \label{eq:class-concept-mean}
\end{equation}
This equality holds even for concepts introduced after class $k$.
Since each class has weight $1/K_t$ and $\mathbf y$ is one-hot, column $k$ of the concept-label cross-moment is the mean concept response for class $k$ scaled by $1/K_t$. Stacking these gives
\begin{equation}
 \mathbf S_{cy}^{(t)}
 =\Emp[\hat{\mathbf c}^{(t)}\mathbf y^\top]
 =\frac{1}{K_t}\mathbf W_c^{(t)}\mathbf P_z^{(t)}.
 \label{eq:head-cross}
\end{equation}

The class means tell us how strongly each concept responds in each class. To update the concept weights jointly, the squared loss also needs the mean products of concept responses across images.
We obtain these products by transforming the stored visual second moment $\mathbf S_{zz}^{(t)}$.
The bias term also requires the mean label $\boldsymbol\mu_y^{(t)}=\mathbf 1/K_t$ and the mean concept response. With $\boldsymbol\mu_z^{(t)}=\mathbf P_z^{(t)}\boldsymbol\mu_y^{(t)}$, the required concept statistics are
\begin{equation}
 \begin{aligned}
 \mathbf S_{cc}^{(t)}
 &=\Emp[\hat{\mathbf c}^{(t)}\hat{\mathbf c}^{(t)\top}]
 =\mathbf W_c^{(t)}\mathbf S_{zz}^{(t)}\mathbf W_c^{(t)\top},\\
 \boldsymbol\mu_c^{(t)}
 &=\mathbf W_c^{(t)}\boldsymbol\mu_z^{(t)} .
 \end{aligned}
 \label{eq:head-moments}
\end{equation}

\begin{proposition}[Full-history equivalence of the classification loss]
\label{prop:head}
Fix the current concept layer $\mathbf W_c^{(t)}$.
For any classifier weights $\mathbf W_y$ and bias $\mathbf b$, evaluating \cref{eq:class-loss} from $\mathbf S_{cy}^{(t)},\mathbf S_{cc}^{(t)},\boldsymbol\mu_c^{(t)},\boldsymbol\mu_y^{(t)}$ yields exactly the same loss as evaluating it on every image seen through phase $t$ with the same class-balanced weights.
\end{proposition}
We can therefore evaluate how changing the weight of a new concept affects the full-history squared classification loss, even when that weight belongs to an earlier class.

\subsection{Graph-Weighted Margin Distortion Penalty}
\label{sec:error}

The classification loss in \cref{eq:class-loss} uses predicted concept scores, so a systematic prediction error can itself become a cue for distinguishing classes.
Suppose two classes have similar VLM reference scores for a new concept, but the concept layer consistently predicts lower scores for one class.
To measure how such errors change the relative scores of classes $k$ and $l$, we evaluate the margin $s_k-s_l$ twice for the same image and classifier, once with predicted concept scores and once with VLM reference scores.

Denote the two margins by $m_{\mathrm{pred}}^{kl}$ and $m_{\mathrm{ref}}^{kl}$, respectively.
Let $\Delta\mathbf w_{kl}=\mathbf w_k-\mathbf w_l$, where $\mathbf w_k^\top$ and $\mathbf w_l^\top$ are the classifier rows for classes $k$ and $l$. Both margins contain the same bias difference $b_k-b_l$, so their difference depends only on the concept error,
\begin{equation}
 \begin{aligned}
 \boldsymbol\delta^{(t)}(x)
 &=\hat{\mathbf c}^{(t)}(x)-\mathbf c_{\mathrm{ref}}^{(t)}(x),\\
 m_{\mathrm{pred}}^{kl}-m_{\mathrm{ref}}^{kl}
 &=\Delta\mathbf w_{kl}^\top\boldsymbol\delta^{(t)}(x).
 \end{aligned}
 \label{eq:concept-error}
\end{equation}
If a concept has equal weights for the two classes, an error in that concept shifts both scores equally and contributes nothing to the margin shift.

We measure distortion for each class pair by its mean squared margin shift over all images seen through phase $t$, using the same class-balanced weights as in \cref{eq:class-loss}.
With $\mathbf R_{\mathrm{err}}^{(t)}=\Emp[\boldsymbol\delta^{(t)}\boldsymbol\delta^{(t)\top}]$, this quantity is
\begin{equation}
 \Emp\!\left[(m_{\mathrm{pred}}^{kl}-m_{\mathrm{ref}}^{kl})^2\right]
 =\Delta\mathbf w_{kl}^\top
 \mathbf R_{\mathrm{err}}^{(t)}\Delta\mathbf w_{kl}.
 \label{eq:margin-identity}
\end{equation}
Using an uncentered moment preserves the contribution of a nonzero mean error. The off-diagonal terms in \cref{eq:margin-identity} account for concept errors whose effects on the margin reinforce or offset one another.
The shared map in \cref{sec:concept-learning} predicts the VLM image embedding with residual $\mathbf W_s^{(t)}\mathbf z-\mathbf u$. Projecting this residual onto the concept text embeddings gives $\boldsymbol\delta^{(t)}=\mathbf E_t^\top(\mathbf W_s^{(t)}\mathbf z-\mathbf u)$.
We can recover $\mathbf R_{\mathrm{err}}^{(t)}$ exactly from $\mathbf S_{zz}^{(t)}$, $\mathbf S_{zu}^{(t)}$, and one additional statistic, the VLM image second moment $\mathbf S_{uu}^{(t)}=\Emp[\mathbf u\mathbf u^\top]$.

We focus the penalty on pairs of classes with similar mean predicted concept responses. At each phase, we center and normalize the class means from \cref{eq:class-concept-mean} and connect each class to its $r$ nearest neighbors, or to all other classes if fewer are available. The resulting undirected graph $\mathcal G_t$ gives closer pairs larger weights $\alpha_{kl}$, normalized to sum to one over its edges when the graph is nonempty.
Each edge selects a pair of classifier rows, while every pair uses the same error moment over all seen classes. Weighting each pair's mean squared margin shift by $\alpha_{kl}$ gives graph-weighted penalty:
\begin{equation}
 \mathcal R_{\mathrm{err}}^{(t)}(\mathbf W_y)
 =\tfrac12\sum_{(k,l)\in\mathcal G_t}
 \alpha_{kl}\Delta\mathbf w_{kl}^\top
 \mathbf R_{\mathrm{err}}^{(t)}\Delta\mathbf w_{kl}.
 \label{eq:graph-risk}
\end{equation}

This penalty discourages classifier weights that amplify concept prediction errors into large margin shifts between neighboring classes. With the concept layer and graph fixed, we learn a single classifier for all seen classes by solving
\begin{equation}
 \begin{aligned}
 \min_{\mathbf W_y,\mathbf b}\quad&
 \mathcal L_{\mathrm{cls}}^{(t)}(\mathbf W_y,\mathbf b)
 +\eta\mathcal R_{\mathrm{err}}^{(t)}(\mathbf W_y)\\
 &+\lambda_1\|\mathbf W_y\|_1
 +\tfrac{\lambda_2}{2}\|\mathbf W_y\|_F^2.
 \end{aligned}
 \label{eq:head-objective}
\end{equation}
Here $\eta\ge0$ controls the penalty on margin shifts and $\lambda_1\ge0$ controls sparsity in the concept weights. The ridge term with $\lambda_2>0$ gives a unique solution for the classifier weights and bias.

\begin{algorithm}[t]
\caption{Training \method{} at phase $t$}
\label{alg:qsm}
\algrenewcommand\algorithmicrequire{\textbf{Input:}}
\algrenewcommand\algorithmicensure{\textbf{Output:}}
\newcommand{\phasecomment}[1]{\Statex \textcolor{black!70}{\(\triangledown\)\enspace\textit{#1}}}
\begin{algorithmic}[1]
\Require $\mathcal D_t$, $\Delta\mathcal C_t$, $\mathbf E_{t-1}$, previous feature statistics
\phasecomment{Update feature statistics}
\State $\mathbf z_i \gets f(x_i),\quad \mathbf u_i \gets g_I(x_i)$ for $(x_i,y_i)\in\mathcal D_t$
\State Update $\mathbf S_{zz},\mathbf S_{zu},\mathbf S_{uu}$ using \cref{eq:moment-update}
\State For new $k$, store $n_k$ and append $\boldsymbol\mu_{z,k},\boldsymbol\mu_{u,k}$ to $\mathbf P_z,\mathbf P_u$
\State Discard the images and per-image features
\phasecomment{Compute the concept layer}
\State $\mathbf E_t \gets [\,\mathbf E_{t-1},\, (g_T(c))_{c\in\Delta\mathcal C_t}\,]$
\State Compute $\mathbf W_s^{(t)}$ by \cref{eq:align-solution} and fix $\mathbf W_c^{(t)} \gets \mathbf E_t^\top\mathbf W_s^{(t)}$
\phasecomment{Update the classifier for all seen classes}
\State Reproject $\mathbf P_z,\mathbf S_{zz}$ using \cref{eq:head-cross,eq:head-moments}
\State Compute factors of $\mathbf R_{\mathrm{err}}^{(t)}$ from $\mathbf W_s^{(t)},\mathbf E_t,\mathbf S_{zz},\mathbf S_{zu},\mathbf S_{uu}$
\State Build and fix $\mathcal G_t$ from class means $\mathbf W_c^{(t)}\boldsymbol\mu_{z,k}$
\State Update $(\mathbf W_y^{(t)},\mathbf b^{(t)})$ by minimizing \cref{eq:head-objective}
\State Carry $\mathbf E_t$ and the updated statistics into the next phase
\Ensure $\mathbf W_c^{(t)},\mathbf W_y^{(t)},\mathbf b^{(t)}$
\end{algorithmic}
\end{algorithm}

\begin{table*}[t]
  \centering
  \caption{Average incremental accuracy (AIA, \%) across six datasets.
    $T$ denotes the number of incremental phases.
    We report mean AIA over the same five random class orders for all methods, with standard deviations shown as subscripts. MuCIL uses 500 samples for replay. \textbf{Bold} marks the highest mean AIA in each column. A dash (-) indicates datasets lacking the per-image concept labels CONCIL requires.}
  \vspace{-1mm}
  \label{tab:recognition-accuracy}
  \scriptsize
  \setlength{\tabcolsep}{1pt}

  \newcommand{\reported}[2]{%
    \mbox{#1\textsubscript{\normalfont\tiny\textcolor[gray]{0.40}{#2}}}}

  \vspace{-1mm}
  \begin{tabular*}{\textwidth}{@{\extracolsep{\fill}}lc*{16}{c}@{}}
    \toprule
    \multirow{2}{*}[-.5ex]{\textbf{Method}}
    & \multirow{2}{*}[-.5ex]{\bfseries\shortstack{Exemplar-\\free}}
    & \multicolumn{1}{c}{\textbf{CIFAR-10}}
    & \multicolumn{3}{c}{\textbf{CIFAR-100}}
    & \multicolumn{3}{c}{\textbf{CUB}}
    & \multicolumn{3}{c}{\textbf{Aircraft}}
    & \multicolumn{3}{c}{\textbf{Places365}}
    & \multicolumn{3}{c}{\textbf{ImageNet}} \\
    \cmidrule(lr){3-3}
    \cmidrule(lr){4-6}
    \cmidrule(lr){7-9}
    \cmidrule(lr){10-12}
    \cmidrule(lr){13-15}
    \cmidrule(lr){16-18}
    &
    & $T{=}5$
    & $T{=}5$ & $T{=}10$ & $T{=}20$
    & $T{=}4$ & $T{=}10$ & $T{=}20$
    & $T{=}5$ & $T{=}10$ & $T{=}20$
    & $T{=}5$ & $T{=}10$ & $T{=}20$
    & $T{=}5$ & $T{=}10$ & $T{=}20$ \\
    \midrule

    ICICLE~\cite{rymarczyk2023icicle} & $\checkmark$
    & \reported{44.3}{1.7}
    & \reported{32.6}{0.8} & \reported{24.7}{0.3} & \reported{17.6}{0.8}
    & \reported{44.4}{0.8} & \reported{30.8}{1.1} & \reported{21.5}{1.0}
    & \reported{21.2}{5.2} & \reported{17.7}{2.8} & \reported{14.2}{0.3}
    & \reported{25.0}{0.6} & \reported{20.3}{0.3} & \reported{13.5}{0.2}
    & \reported{21.3}{0.5} & \reported{17.8}{0.3} & \reported{12.4}{0.1} \\
    \addlinespace[2pt]

    IN2~\cite{yang2024conceptdriven} & $\checkmark$
    & \reported{46.7}{0.2}
    & \reported{41.2}{0.3} & \reported{28.0}{0.3} & \reported{17.6}{0.2}
    & \reported{36.0}{0.5} & \reported{20.4}{0.3} & \reported{13.9}{0.1}
    & \reported{10.5}{0.4} & \reported{10.8}{0.1} & \reported{10.6}{0.4}
    & \reported{29.1}{0.5} & \reported{22.5}{0.6} & \reported{15.9}{0.1}
    & \reported{22.5}{0.5} & \reported{17.5}{0.1} & \reported{12.9}{0.1} \\
    \addlinespace[2pt]

    CONCIL~\cite{lai2025concil} & $\checkmark$
    & --
    & -- & -- & --
    & \reported{55.2}{0.3} & \reported{51.7}{0.2} & \reported{48.5}{1.6}
    & -- & -- & --
    & -- & -- & --
    & -- & -- & -- \\
    \addlinespace[2pt]

    CI-CBM~\cite{javadi2026cicbm} & $\checkmark$
    & \reported{80.3}{0.3}
    & \reported{68.6}{0.1} & \reported{68.9}{0.1} & \reported{67.9}{0.1}
    & \reported{67.6}{0.3} & \reported{69.0}{0.1} & \reported{68.5}{0.3}
    & \reported{30.4}{0.3} & \reported{29.5}{0.9} & \reported{29.2}{0.7}
    & \reported{48.5}{0.2} & \reported{50.7}{1.0} & \reported{49.4}{0.1}
    & \reported{32.7}{0.6} & \reported{31.9}{0.2} & \reported{31.9}{0.1} \\
    \addlinespace[2pt]

    MuCIL~\cite{agrawal2025mucil} & $\times$
    & \reported{66.3}{1.2}
    & \reported{54.4}{0.5} & \reported{47.9}{0.3} & \reported{42.6}{1.1}
    & \reported{66.2}{0.5} & \reported{65.0}{0.4} & \reported{62.3}{0.4}
    & \reported{47.6}{0.2} & \reported{45.9}{0.5} & \reported{43.0}{0.4}
    & \reported{31.5}{0.7} & \reported{24.6}{0.2} & \reported{17.7}{0.3}
    & \reported{29.6}{0.1} & \reported{23.1}{0.5} & \reported{17.9}{0.9} \\
    \midrule

    \textbf{RetroCBM} & $\checkmark$
    & \reported{\textbf{89.6}}{2.4}
    & \reported{\textbf{71.9}}{0.8} & \reported{\textbf{73.6}}{0.7} & \reported{\textbf{74.6}}{0.5}
    & \reported{\textbf{72.4}}{0.2} & \reported{\textbf{74.2}}{0.2} & \reported{\textbf{75.0}}{0.1}
    & \reported{\textbf{50.0}}{2.8} & \reported{\textbf{51.1}}{2.4} & \reported{\textbf{52.3}}{3.0}
    & \reported{\textbf{55.5}}{1.4} & \reported{\textbf{57.1}}{1.4} & \reported{\textbf{58.3}}{1.3}
    & \reported{\textbf{38.0}}{1.2} & \reported{\textbf{39.8}}{1.4} & \reported{\textbf{40.9}}{1.5} \\
    \bottomrule
  \end{tabular*}
  \vspace{-2mm}
\end{table*}

\subsection{Class-Incremental Training and Inference}
\label{sec:optimization}
At each phase, we update the class-balanced feature statistics from the incoming images and compute the concept layer for all current concepts. With this layer fixed, we reproject the stored statistics to update the classifier for all seen classes without revisiting earlier images (\cref{alg:qsm}).
At inference, the frozen visual encoder $f$ and the precomputed concept layer $\mathbf W_c^{(t)}$ give the concept responses for a test image $x$.
Each concept $j$ contributes $[\mathbf W_y^{(t)}]_{kj}\hat c_j^{(t)}(x)$ to the score of class $k$ in \cref{eq:model}. We sum these contributions, add the class bias $b_k^{(t)}$, and predict the highest-scoring class.

\paragraph{Complexity Analysis.} The three global moments have fixed dimensions determined by the encoders. Each new class adds only $d+m+1$ values for two mean vectors and a count, regardless of how many images it contains. To summarize, these feature statistics occupy $O(d^2+dm+m^2+K_t(d+m))$ numbers. We also retain one $m$-dimensional text embedding per concept, so for fixed encoders the storage carried across phases grows linearly with the class and concept counts. After ten phases on CUB, the learned parameters used at inference occupy about 2~MiB (\cref{fig:cub-comparison}, left).

\section{Experiments}
\label{sec:exp}
In this section, we first describe the experimental setup and compare our \method{} with baselines on six datasets. We then report ablation studies, sensitivity to regularization parameters, and inference storage costs. Finally, we conduct a qualitative case study to illustrate its interpretability.

\subsection{Experimental Setup}
\label{sec:exp-setup}
\paragraph{Datasets and Protocols.}
We evaluate \method{} on CIFAR-10 and CIFAR-100~\cite{krizhevsky2009cifar}, CUB~\cite{wah2011cub}, Aircraft~\cite{maji2013aircraft}, Places365~\cite{zhou2018places}, and ImageNet~\cite{russakovsky2015imagenet}. We follow the official data splits for training and testing. Each phase introduces new classes and their associated concepts from a vocabulary shared with CI-CBM. We reuse existing concept lists and supplement them as needed with candidates generated from class names by GPT-5 and filtered with CLIP~\cite{radford2021clip}. After each phase, we discard its training images and per-image features. \Cref{tab:recognition-accuracy} lists the phase counts for datasets.

\paragraph{Baselines.}
We compare our \method{} with existing state-of-the-art interpretable methods designed for class-incremental learning. These include ICICLE~\cite{rymarczyk2023icicle}, IN2~\cite{yang2024conceptdriven}, CONCIL~\cite{lai2025concil}, CI-CBM~\cite{javadi2026cicbm}, and MuCIL~\cite{agrawal2025mucil}.

\paragraph{Implementation Details.}
\method{} and CI-CBM use the same 512-dimensional features from a frozen ResNet-18~\cite{he2016deep}, the same SigLIP ViT-L/16-384 teacher~\cite{zhai2023siglip}, and the same concept vocabulary.
We tune each method on the validation set for each dataset.

\paragraph{Evaluation Metrics.}
We report \emph{average incremental accuracy} (AIA), $\mathrm{AIA}=\frac{1}{T}\sum_{t=1}^{T}a_t$, where $a_t$ is the top-1 accuracy on images from all classes seen through phase $t$. \emph{Old-class accuracy} measures top-1 accuracy on images from earlier classes, with all seen classes as candidates. For each late-arriving concept, we use the VLM to identify the 10\% of evaluation images from earlier classes that best match the concept description. \emph{Old-image recall} is the fraction of these images that the concept predictor also ranks in its top 10\%.

\subsection{Main Results}
\label{sec:exp-main}
\Cref{tab:recognition-accuracy} reports average incremental accuracy (AIA) across six datasets. We make three observations.
(1) Averaged across all 16 settings, \method{} achieves an 18.1\% relative gain in AIA over CI-CBM, the strongest exemplar-free baseline in every setting. This result shows that \method{} can use stored feature statistics to update its expanding concept bottleneck without revisiting earlier images and still achieve higher average accuracy than CI-CBM.
(2) \method{} outperforms CI-CBM at every phase count on both CUB and Aircraft, with the largest gain of 23.1 percentage points on Aircraft at 20 phases. These results show that the expanding concept bottleneck remains effective for fine-grained distinctions among bird species and aircraft models.
(3) On four of the five datasets tested at multiple phase counts, \textbf{the AIA gap in favor of \method{} widens as the same classes are split across more phases.} With fewer new classes per phase, supervision from earlier classes becomes more important for learning shared concepts. \method{} incorporates this supervision through stored statistics and updates old-class weights, which may explain its larger gains in these settings.

\subsection{Ablation Studies}
\label{sec:exp-ablation}
\begin{figure}[t]
    \centering
    \includegraphics[width=\columnwidth]{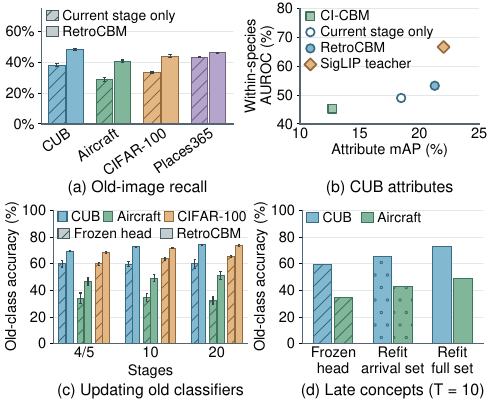}
    \caption{\textbf{Ablation Study.}
    (a) Old-image recall for new concepts learned with or without historical statistics.
    (b) Concept predictions evaluated against human attribute labels on CUB.
    (c) Old-class accuracy with frozen or updated classifiers.
    (d) Old-class accuracy when classifiers are refitted with or without late-arriving concepts.}
    \label{fig:retro-ablations}
\end{figure}

\Cref{fig:retro-ablations} shows how historical statistics help learn new concepts and how old-class classifiers use them. We make three observations.
(1) \textbf{Historical statistics improve recognition of late-arriving concepts.}
When a new concept arrives, we fit its predictor either on current-phase images alone or with additional statistics from earlier classes.
Using these statistics improves recall of the VLM's top-ranked old images by 3.02 to 11.72 percentage points across four datasets (\cref{fig:retro-ablations}(a)).
We also evaluate predictions against human attribute labels on CUB (\cref{fig:retro-ablations}(b)).
Attribute mAP rises from 18.47\% to 21.28\%, and within-species AUROC rises from 48.98\% to 53.22\%, indicating better attribute recognition even among birds of the same species.
(2) \textbf{Refitting old-class classifiers improves old-class accuracy.}
With the concept predictor fixed, we either freeze the old-class weights or refit them in the expanded concept space through moment reprojection (\cref{fig:retro-ablations}(c)).
Refitting improves mean old-class accuracy in all nine settings on CUB, Aircraft, and CIFAR-100, with gains of 8.05 to 18.53 percentage points.
(3) \textbf{Late-arriving concepts provide an additional gain beyond refitting.}
We refit the old-class classifiers in both conditions, allowing each class to use either the concepts available at its arrival or all concepts seen so far (\cref{fig:retro-ablations}(d)).
Giving old classes access to later concepts adds 7.39 percentage points to old-class accuracy on CUB and 6.60 points on Aircraft in the ten-phase setting.

\begin{figure}[!t]
    \centering
    \includegraphics[width=\columnwidth]{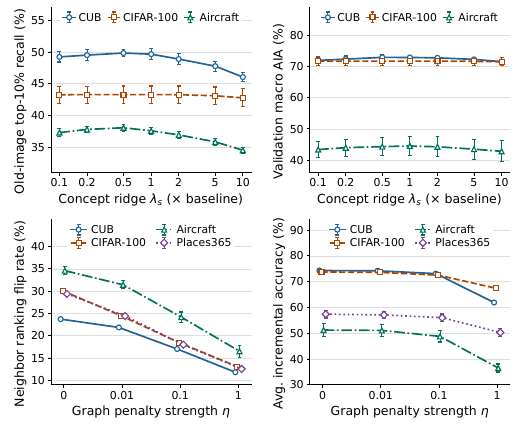}
    \caption{\textbf{Sensitivity to Regularization.} The top row shows old-image recall (left) and validation macro AIA (right) as we vary the concept ridge coefficient $\lambda_s$ relative to its baseline for each dataset. The bottom row shows the ranking flip rate for neighboring classes (left) and AIA (right) as we vary the graph penalty $\eta$.}
    \label{fig:exp-regularization}
\end{figure}

\begin{figure*}[t]
    \centering
    \includegraphics[width=\textwidth]{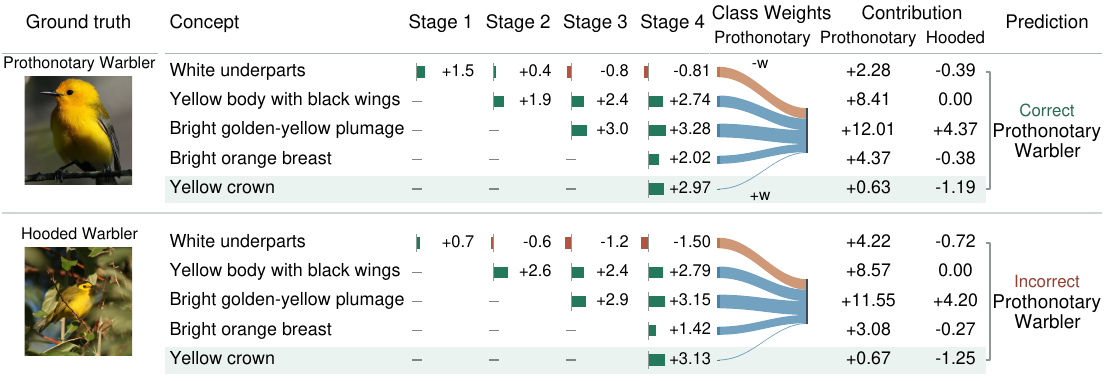}
    \vspace{-4mm}
    \caption{\textbf{Tracing a Late-Arriving Concept on CUB.} Rows show selected concept responses across phases, with dashes before each concept is introduced. Blue and orange bands denote positive and negative weights for Prothonotary Warbler, respectively. The last two numeric columns show each concept's contribution to the two species' scores in phase four.}
    \vspace{-2mm}
    \label{fig:exp-case}
\end{figure*}

\begin{figure}[t]
    \centering
    \includegraphics[width=.49\columnwidth]{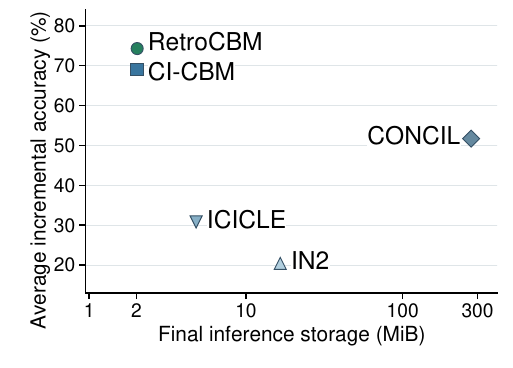}\hfill
    \includegraphics[width=.49\columnwidth]{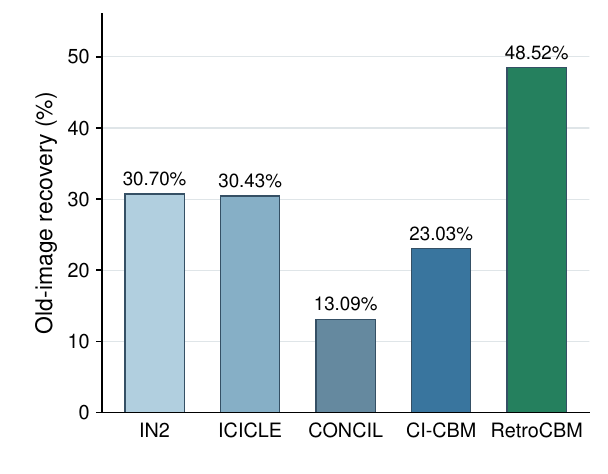}
    \vspace{-1mm}
    \caption{\textbf{Inference Storage and Old-Image Recall.} On CUB with ten phases, we compare AIA versus final inference storage (left) and old-image recall for late-arriving concepts (right).}
    \vspace{-1mm}
    \label{fig:cub-comparison}
\end{figure}

\subsection{Additional Quantitative Analyses}
\label{sec:exp-sensitivity}
\begin{table}[t]
  \centering
  \caption{\textbf{Comparison with Full Rehearsal.}
  We report relative errors in concept weights and maximum absolute differences in concept scores, classifier objective values, and sub-gradients.}
  \label{tab:full-history-equivalence}
  \small
  \setlength{\tabcolsep}{3pt}
  \renewcommand{\arraystretch}{1.12}
  \resizebox{\columnwidth}{!}{%
    \begin{tabular}{@{}lcccc@{}}
      \toprule
      \multirow{3}{*}[-.5ex]{Dataset}
        & \multicolumn{2}{c}{Concept predictor}
        & \multicolumn{2}{c}{Sparse classifier} \\
      \cmidrule(lr){2-3}\cmidrule(l){4-5}
        & Concept Weight & Concept Score & Objective  & Sub-gradient \\
        & Error          & Error         & Difference & Difference  \\
      \midrule
      CUB       & $2.51\times10^{-13}$ & $1.34\times10^{-14}$
                & $4.62\times10^{-14}$ & $1.90\times10^{-15}$ \\
      CIFAR-100 & $3.43\times10^{-13}$ & $2.66\times10^{-14}$
                & $1.36\times10^{-12}$ & $5.22\times10^{-15}$ \\
      Aircraft  & $2.96\times10^{-13}$ & $6.88\times10^{-15}$
                & $1.63\times10^{-13}$ & $2.91\times10^{-15}$ \\
      \bottomrule
    \end{tabular}%
  }
\end{table}

\paragraph{Sensitivity to Regularization.}
\Cref{fig:exp-regularization} shows how the concept regression coefficient $\lambda_s$ and the graph penalty $\eta$ affect concept predictions and classification accuracy. On CUB and Aircraft, old-image recall is more sensitive to $\lambda_s$ than validation macro AIA (top row). Recall peaks near half the baseline value of $\lambda_s$ and declines as regularization grows stronger, while recall on CIFAR-100 remains nearly constant over the same range. Thus, even when classification accuracy changes little, the choice of $\lambda_s$ can affect how well a new concept retrieves earlier-class images that the VLM ranks highly. The graph penalty reduces how often neighboring classes exchange rank when the classifier uses predicted concept scores in place of VLM scores (bottom row). Across all four datasets, a weak penalty ($\eta=0.01$) reduces these ranking flips while keeping AIA close to its value without penalty. A stronger penalty reduces flips further at a cost to accuracy. On CUB, increasing $\eta$ from $0$ to $1$ roughly halves the flip rate but lowers AIA by about 12 percentage points.

\paragraph{Comparison with Full Rehearsal.}
To test what the stored statistics preserve, we compare \method{} with an offline reference that uses individual features from all training images. We perform this comparison after the final phase of the ten-phase experiments on CUB, CIFAR-100, and Aircraft (\cref{tab:full-history-equivalence}). The two approaches yield numerically identical concept weights and scores on the same test images. For the classifier, they also give the same objective value and subgradient at fixed weights, up to numerical precision. The largest discrepancy is $1.36\times10^{-12}$. These results confirm that \textbf{the stored statistics preserve the contribution of earlier images both to learning late-arriving concepts and to evaluating the classifier objective in concept space.}

\paragraph{Inference Storage and Old-Image Recall.}
On CUB with ten phases, \method{} achieves the highest AIA while storing about 2~MiB of learned parameters for inference (\cref{fig:cub-comparison}, left). ICICLE, IN2, and CONCIL all use more storage and achieve lower AIA. Even CONCIL, the most accurate of these three, uses over 100 times as much storage as \method{}. CI-CBM matches the storage cost of \method{} but trails it by 5.2 percentage points in AIA. \textbf{At this same storage cost, \method{} also more than doubles old-image recall}: it retrieves 48.52\% of the old-class images that the VLM ranks highest for late-arriving concepts, compared with 23.03\% for CI-CBM (\cref{fig:cub-comparison}, right).

\subsection{Case Study}
\label{sec:exp-case}
\Cref{fig:exp-case} traces how the concept \emph{yellow crown}, introduced in phase four, affects the predictions for two CUB images. In the top image, the Prothonotary Warbler has a visibly yellow crown, and \method{} assigns the concept a response of \(+2.97\). This response adds \(0.63\) to the Prothonotary Warbler score and subtracts \(1.19\) from the Hooded Warbler score. The late-arriving concept thus contributes \(1.82\) points to the margin in favor of the correctly predicted Prothonotary Warbler.
In the bottom image, \method{} assigns an even higher \emph{yellow crown} response (\(+3.13\)) to a Hooded Warbler. With the same classifier weights, this response raises the Prothonotary Warbler score by \(0.67\) and lowers the Hooded Warbler score by \(1.25\). The model again predicts Prothonotary Warbler. \textbf{In this case, the same late-arriving concept favors the incorrect class by \(1.92\) points.}

\section{Conclusion}
\label{sec:conclusion}
In this paper, we proposed \method{}, a concept bottleneck model for exemplar-free class-incremental learning. \method{} uses stored feature statistics to learn newly introduced concepts from earlier classes after their training images have been discarded. By reprojecting the same statistics into the expanded concept space, the classifier learns how to combine new and existing concepts for each earlier class. A graph-weighted penalty limits the influence of concept prediction errors on the score margins between similar classes. Across six datasets, \method{} outperforms state-of-the-art interpretable methods in all 16 settings, with an 18.1\% relative gain in mean AIA over the strongest exemplar-free baseline. Our CUB case study shows how the late-arriving concept \emph{yellow crown} supports a correct prediction for one bird and reinforces a misclassification for another.

{\small
\bibliographystyle{ieeenat_fullname}
\bibliography{refs}
}
\end{document}